\documentclass[final,10pt,authoryear]{elsarticle}
\usepackage{amssymb}

\usepackage{rotating}

\usepackage{booktabs}
\usepackage{makecell}
\usepackage{xcolor}

\usepackage{hyperref} 

\usepackage{footmisc}

\usepackage{enumerate, enumitem} 
\usepackage{hhline, threeparttable} 
\usepackage{array} 

\usepackage{amssymb,mathtools} 
\usepackage{multirow}
\usepackage{amsmath}

\usepackage{mathtools, nccmath}

\journal{arXiv}

\begin{document}

\begin{frontmatter}



\title{Unigram-Normalized Perplexity as a Language Model Performance Measure with Different Vocabulary Sizes}


\author[mymainaddress]{Jihyeon Roh}

\author[mythirdaddress]{Sang-Hoon Oh}

\author[mymainaddress]{Soo-Young Lee\corref{mycorrespondingauthor}}
\cortext[mycorrespondingauthor]{Corresponding author}
\ead{sy-lee@kaist.ac.kr}

\address[mymainaddress]{School of Electrical Engineering, Korea Advanced Institute of Science and Technology, Daejeon, Korea}
\address[mythirdaddress]{Division of Information and Communication Convergence Engineering, Mokwon University, Daejeon, Korea}

\begin{abstract}
Although Perplexity is a widely used performance metric for language models, the values are highly dependent upon the number of words in the corpus and is useful to compare performance of the same corpus only. In this paper, we propose a new metric that can be used to evaluate language model performance with different vocabulary sizes. The proposed unigram-normalized Perplexity actually presents the performance improvement of the language models from that of simple unigram model, and is robust on the vocabulary size. Both theoretical analysis and computational experiments are reported.

\end{abstract}

\begin{keyword}
Language models\sep natural language processing\sep performance measure \sep vocabulary size 
\end{keyword}

\end{frontmatter}

\section{Introduction}
Language model (LM) is a core elements in natural language processing (NLP) applications, e.g., language modeling \citep{bengio2003neural,mikolov2011extensions}, machine translation~\citep{cho2014learning}, speech recognition~\citep{amodei2016deep}, and dialogue generation. LMs determine the probability of word sequences and provide a metric for the probability of generated sequences. To achieve good performance, LMs must accurately capture the relationship among words and phrases in word sequences. 

The popular metric of LM performance is Perplexity, which is based on the likelihood of word sequences. Since the probability of words itself is highly varying for different corpus of different vocabulary size, the likelihood of word sequences is also very sensitive on corpus. For an example, the Perplexity value 10 may be very good for some corpus, but very poor for another. Therefore, to compare LM performances of different corpus, we propose a new metric, i.e., unigram-normalized Perplexity ($PPLu$), which is independent of the vocabulary size.

\section{Unigram-Normalized Perplexity (\textit{PPLu})}

\subsection{Definition}\label{perplexityppl}
To evaluate the performance of LMs, Perplexity (\textit{PPL}) has been used as a popular metric. The \textit{PPL} of a word sequence is computed as the geometric average of the inverse probability of the words: 
\begin{equation}
PPL = P(w_1,...,w_T)^{-\frac{1}{T}} = \left(\prod_{t=1}^{T} {P(w_t | w_{1:t-1})}\right)^{-\frac{1}{T}},
\label{eq_PPL}
\end{equation}
where $T$ is the length of the sequence and $w_t$ is the word at time $t$. An LM that produces a higher probability for test sequences can obtain a lower \textit{PPL}, i.e., better performance. This \textit{PPL} metric is suitable for comparing LMs using the same vocabulary. However, \textit{PPL} may not be suitable for comparing LMs using different vocabularies because a larger vocabulary size tends to result in lower word probabilities and thus a higher \textit{PPL}, i.e., worse performance. 

To overcome the limitations of the perplexity, we adopt the basic idea of normalizing the word probability with respect to a quantity containing the vocabulary size.
We apply a unigram probability that is calculated from the word occurrence
as a normalization factor for the perplexity. The unigram probability from the unigram LM is computed as $P_{uni}(w_t=v_k)=Count(v_k)/\sum_{k'=1}^{|V|} Count(v_{k'})$, where $Count(v_k)$ is the number of occurrences of word $v_k$ in the corpus. 

Our proposed metric is obtained by normalizing the perplexity with this unigram probability. The proposed ``PPL normalized with unigram'' (\textit{PPLu}) is defined as
\begin{equation}
\begin{aligned} 
PPLu & = \left(\prod_{t=1}^T \frac{P(w_t|w_{1:t-1})}{P(w_t)}\right)^{-\frac{1}{T}}
\end{aligned}
\end{equation}
This metric shows the likelihood improvement of a context-dependent LM from unigram LM without the context information, and enables us to evaluate the effectiveness of an LM.

\subsection{\textit{PPLu} in Terms of Mutual Information} \label{interpret_pplu}
In this section, we describe the \textit{PPLu} by the form of mutual information as the additional interpretation of \textit{PPLu}. 
From the definition of \textit{PPLu}, it can be computed as
\begin{equation}
\begin{aligned} 
PPLu & = \left(\prod_{t=1}^T \frac{P(w_t|w_{1:t-1})}{P(w_t)}\right)^{-\frac{1}{T}}
 & = \left(\prod_{t=1}^T \frac{P(w_t,w_{1:t-1})}{P(w_t)P(w_{1:t-1})}\right)^{-\frac{1}{T}} .
\end{aligned}
\end{equation}
By taking the logarithm, we obtain 
\begin{equation}
\begin{aligned} 
\log {PPLu} = -\frac{1}{T}\left(\sum_{t=1}^T \log  \frac{P(w_t,w_{1:t-1})}{P(w_t)P(w_{1:t-1})}\right) .
\end{aligned}
\end{equation}

Let $w_{1:t}$ be random variables, the expectation of $\log PPLu$ given the random variables,
\begin{equation}
    \mathbb{E}[\log PPLu]=-\frac{1}{T}\sum_{t=1}^{T} \mathbb{E}\left[\log \frac{P(w_t,w_{1:t-1})}{P(w_t)P(w_{1:t-1})}\right].
\end{equation}
This can be shown as the mutual information between $w_t$ and $w_{1:t-1}$:
\begin{equation}
\begin{aligned} 
&\mathbb{E}\left[\log \frac{P(w_t,w_{1:t-1})}{P(w_t)P(w_{1:t-1})}\right]
&=\int_{w_{1:t}} P(w_t,w_{1:t-1}) \log \frac{P(w_t,w_{1:t-1})}{P(w_t)P(w_{1:t-1})}dw_t dw_{t-1}...dw_1
\\ & = I(w_t;w_{1:t-1}) .
\end{aligned}
\end{equation}
Here, $I(w_t;w_{1:t-1})$ denotes the mutual information between $w_t$ and $w_{1:t-1}$. This measures the amount of shared information between the word $w_t$ and the context $w_{1:t-1}$. $I(w_t;w_{1:t-1})\geq 0$, and $I(w_t;w_{1:t-1})=0$ when $w_t$ and $w_{1:t-1}$ are independent. Thus, we can argue that $\mathbb{E}[\log PPLu]$ depends only on the dependency between $w_t$ and $w_{1:t-1}$.

\subsection{Proof of the consistency of \textit{PPLu} with different vocabulary sizes} \label{proof_pplu}
In this subsection, we attempt to prove the consistency of \textit{PPLu} according to the vocabulary size.

Let us consider a single word $v_{ab}$ to be split into two words $v_a$ and $v_b$. The vocabulary size of the original document is $|V|$, and the vocabulary size after the word is split is $|V|+1$. Then, $P(v_{ab})=P(v_a)+P(v_b)$ and
\begin{equation}
\begin{aligned} 
& P(\textbf{s}=(w_1,…,w_i=v_{ab},…,w_N ))
\\& =P(\textbf{s}=(w_1,…,w_i=v_a,…,w_N )) +P(s=(w_1,…,
& \;\; w_i=v_b,…,w_N )) ,
\end{aligned} 
\label{eqS_prob1}
\end{equation} 
where $N$ is the number of tokens in a sentence $\textbf{s}$. For simplicity, we assume that only one instance of $v_{ab}$ is included in a sentence. However, this approach may be easily extended toward multiple instances of split words in a sentence.

\subsubsection{Perplexity (\textit{PPL}) change after word splitting}
\begin{enumerate}[]
\item{	Baseline with vocabulary size $|V|$:}

The perplexity with the original sentence \textbf{s} is computed as
\begin{equation}
\begin{aligned} 
PPL^B &=\exp {\left( -\frac{1}{N}\log P(\textbf{s}) \right)}
\\&=\exp{\left(-\frac{1}{N}\log P(w_1,...,w_N) \right)};
\label{PPLB}
\end{aligned}
\end{equation}
Then the logarithm of $PPL^B$ is computed as
\begin{equation}
\begin{aligned} 
\log{PPL^B}={-\frac{1}{N}\log P(w_1,...,w_N)}.
\end{aligned}
\end{equation}

\item{	Word splitting with vocabulary size $|V|+1$:}

\indent In this case, the word $v_{ab}$ is split into two words $v_{a}$ and  $v_{b}$. The vocabulary size is increased by $1$ compared to the original size. Since $v_a$ and $v_b$ are originally located at the same position and all other words are the same, the perplexity of a sentence $\textbf{s}$ with $v_{ab}$ in the split case is computed as  
\begin{equation}
\begin{aligned} 
\log{PPL^S} &={-\frac{1}{N}\log P(w_1,...,w_i=v_{a} ~ or ~ v_{b},...,w_N)}
\\& > \log{PPL^B} \\&={-\frac{1}{N} 
\log P(w_1 ,...,w_i=v_{ab},...,w_N) } .
\end{aligned}
\end{equation}
For all the other sentences without $v_{ab}$, the probabilities are unchanged. Therefore, $PPL^S$ is larger than $PPL^B$ when the vocabulary size is increased. If the words $v_a$ and $v_b$ have identical meaning, the LM performance in both cases should be equal. Any difference between the two cases arises only from the increase of the vocabulary size.
\end{enumerate}

\subsubsection{Same (\textit{PPLu}) after splitting}
Similarly, we only need to compare \textit{PPLu} for sentences with $v_{ab}$.
\begin{enumerate}[]
\item{Baseline with vocabulary size $|V|$:}

If a sentence includes $v_{ab}$, then 
\begin{equation}
\begin{aligned} 
     \log{PPLu^B}=&-\frac{1}{N}[\log P(w_1,...,w_i=v_{ab},...,w_N) \\ 
     &-\{\sum_{n\neq i}^N \log P(w_n) + \log P(w_i=v_{ab})\}] .
\end{aligned}
\end{equation}

\item{Word splitting with vocabulary size $|V|+1$:}

If a sentence includes either $v_a$ or $v_b$, then
\begin{equation}
\begin{aligned} 
     \log{PPLu^S}=-\frac{1}{N}[\log P(w_1,...,w_i=(v_a \, or \, v_b),... ,w_N) \\ -\{\sum_{n\neq i}^N \log P(w_n) + \log P(w_i=(v_a \, or \, v_b))\}] .
\end{aligned}
\end{equation}Although the first term of $\log PPLu^S$ is larger than that of $\log PPLu^B$, the second terms have the same tendency. Therefore, the difference due to the change in the vocabulary size may become smaller.

If the word $v_{ab}$ actually has only one meaning and we randomly assign $v_a$ or $v_b$ instead of $v_{ab}$ with probabilities $\beta$ and (1-$\beta$), respectively, then 
\begin{equation}
\begin{aligned}
P(v_a) = \beta P(v_{ab}), 
P(v_b) = (1- \beta ) P(v_{ab}), \\
P(…,w_i=v_a,…) = \beta P(…,w_i=v_{ab},…), \\
P(…,w_i=v_b,…) = (1- \beta ) P(…,w_i=v_{ab},…).
\end{aligned}
\end{equation}
Therefore, for a sentence with $v_a$,
\begin{equation}
\begin{aligned}
\log{PPLu^S}=-\frac{1}{N}[\log \beta {P(...,w_i=v_{ab},...)} \\
- \{ \sum_{n\neq i}^N \log P(w_n)  + \log \beta P(w_i=v_{ab}) \}] \\
= \log{PPLu^B} .
\end{aligned}
\end{equation}
The same is true for sentences with $v_b$.
In this case, \textit{PPLu} is invariant with the split of a word even with the increased vocabulary size.

If the word $v_{ab}$ has multiple senses and the word is split into multiple words based on the meaning, the split is no longer random and $\beta$, the ratio of $P(...,w_i=v_{a},...)$ and $P(...,w_i=v_{ab},...)$, is no longer a fixed constant and depends upon the other words in the sentence. Therefore, the \textit{PPLu} with or without the split become different. Since \textit{PPLu} is invariant with the vocabulary size, the difference in value of \textit{PPLu} purely comes from the ability of LM.
\end{enumerate}

\begin{figure*}[t!]
 \centering
  \includegraphics[width=13cm]{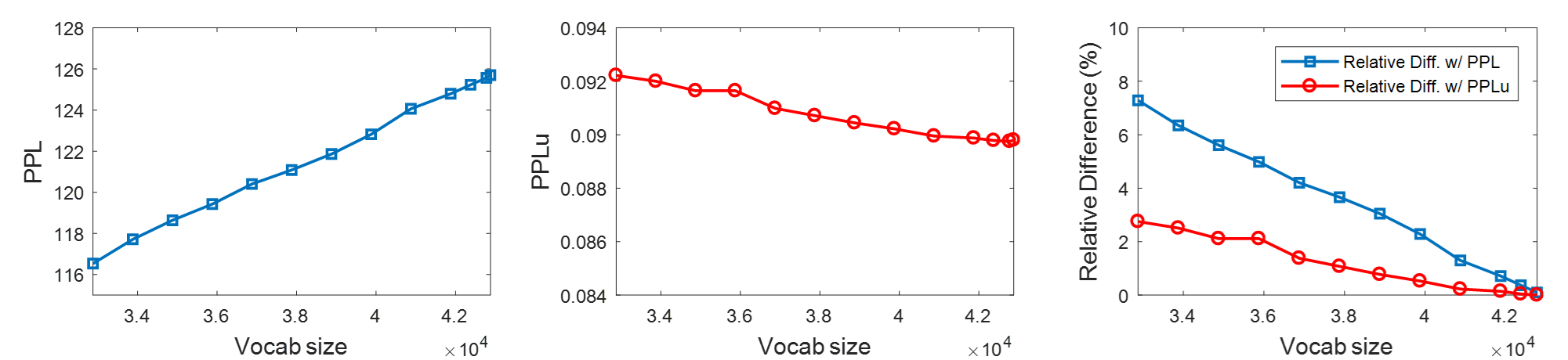}
    \caption[]{{Performance on the Text8 test set for different vocabulary sizes using (a) \textit{PPL} and (b) \textit{PPLu}. (c) Relative differences with respect to \textit{PPL} (blue line) and \textit{PPLu} (red line) between the full-vocabulary model and the models with a reduced vocabulary size.}} \label{FigS1_PPLu}
\end{figure*}

\section{Experimental Results}
We present an additional empirical simulation to show that our \textit{PPLu} metric is less sensitive to different vocabulary sizes than the vanilla \textit{PPL} metric is. Conceptually, our \textit{PPLu} is computed as a ratio between two probabilities, both of which are affected by the vocabulary size, thus enabling us to cancel out the effect of the vocabulary size. In Fig.~\ref{FigS1_PPLu}, we further verify this explanation by evaluating the same model with both the \textit{PPL} and \textit{PPLu} metrics under different vocabularies. Specifically, this model was trained on Text8 with the full vocabulary of 44k words and was tested by sequentially discarding words that rarely appear. The relative difference was computed as
\begin{equation}
 \frac{(|Metric(Vocab_{Full})-Metric(Vocab_{Reduced})|)}{Metric(Vocab_{Full})} \times 100 (\%),
\end{equation}
\noindent where $Metric(Vocab_{Full})$ denotes either \textit{PPL} or \textit{PPLu} measured with the full vocabulary, and $Metric(Vocab_{Reduced})$ denotes the same metric measured with a reduced vocabulary size. This figure confirms that our\textit{PPLu} is less affected by varying vocabulary sizes than \textit{PPL} is.

\begin{table*}[t]
  \centering
  \caption{Performance on the PTB\textsubscript{sub} test set using $PPL_{all}$ and \textit{PPLu}.  The best results among the models are shown in bold text.}
  \label{Table_supp_PPLall}
    \begin{tabular}{lcc}
    \toprule
    \textbf{Model} & \multicolumn{1}{c}{\textit{PPL\textsubscript{all}}} & \multicolumn{1}{c}{\textit{PPLu}} \\
    \midrule
    SSLM--LSTM--W200 & 142.4 & 0.1527 \\
    SSLM--LSTM--W650 & 135.6 & 0.1317 \\
    \midrule
    MSLM--LSTM--W200 (K=2) (ours) & 135.3 & 0.1302 \\
    MSLM--LSTM--W650 (K=2) (ours) & \textbf{128.9} & \textbf{0.1195} \\
    \bottomrule
    \end{tabular}%
  \label{tab:addlabel}%
\end{table*}%

\begin{table*}[htbp]
  \centering
  \caption{Top-3 sentences in the PTB test set sorted by (a) \textit{PPL} and (b) \textit{PPLu}}
  \label{Table_PPLuSentence}
    \begin{tabular}{rr}
    \toprule
    Rank  & \makecell{(a) Sentences sorted by \textit{PPL}} \\
    \midrule
    1     & \makecell{the following were among friday 's offerings and pricings in\\ the u.s. and non-u.s. capital markets with terms and syndicate\\ manager as compiled by dow jones capital markets report} \\
    \midrule
    2     & \makecell{N N N to N days N N N to N days N N N to N days N N N to N days\\ N N N to N days N N N to N days N N N to N days N N N to N days}  \\
    \midrule
    3     & \makecell{N N N to N N N one month N N N to N N N two months\\N N N to N N N three months N N N to N N N four months\\N N N to N N N five months N N N to N N N six months}\\
    \midrule
    \midrule
      & \makecell{(b) Sentences sorted by \textit{PPLu}} \\
    \midrule
    1     & \makecell{the following were among friday 's offerings and pricings in\\ the u.s. and non-u.s. capital markets with terms and syndicate\\ manager as compiled by dow jones capital markets report} \\
    \midrule
    2    & \makecell{source fulton prebon u.s.a inc} \\
    \midrule
    3     & \makecell{ the key u.s. and foreign annual interest rates below are a guide\\to  general levels but do n't always represent actual transactions} \\
    \bottomrule
    \end{tabular}%
  \label{tab:addlabel}%
\end{table*}%

Similar to the original \textit{PPL}, our \textit{PPLu} also contains a conditional probability term and thus offers evidence about the generation quality. Moreover, \textit{PPLu} additionally contains a unigram probability term, which allows \textit{PPLu} to evaluate LMs more accurately than \textit{PPL} does. Specifically, even if an LM fails to capture word relationships, it may achieve a good \textit{PPL} by simply assigning high probabilities to words that frequently appear (e.g., unknown tokens). This case can be corrected with our \textit{PPLu}, which considers the word frequencies via unigram probabilities. 
The following results qualitatively show that better \textit{PPLu} values are assigned to sentences with more natural contexts.
Table~\ref{Table_PPLuSentence} shows the top-3 sentences in the PTB test set sorted by their \textit{PPL} or \textit{PPLu} values as obtained from the baseline LM. The results based on \textit{PPL} and \textit{PPLu} mainly differ in 1) the naturalness of the sentences and 2) the ranking of sentences containing frequently occurring words.

Regarding 1), the \textit{PPL} value tends to be lower as more frequent words appear in the sentence. Moreover, sentences with lower \textit{PPLu} values generally have a more natural form than sentences with lower \textit{PPL} values. These results show how \textit{PPLu} reflects the actual performance of language modeling. Regarding 2), in a similar way, we compared the ranking of sentences in terms of \textit{PPL} and \textit{PPLu} given frequently occurring words. Sentences composed of frequently occurring words rank high in terms of \textit{PPL} and low in terms of \textit{PPLu}. For example, the sentence that is ranked 2-nd in terms of its \textit{PPL} value is ranked 166-th in terms of its \textit{PPLu} value.

\section{Conclusion} \label{Conclusion}
In this paper, we proposed a new evaluation metric for language model performance, i.e., unigram-normalized PPL (\textit{PPLu}), and provided both theoretical and experimental results. 
The \textit{PPLu} denotes the likelihood improvement of word sequences from that of unigram model.
The new metric is independent from the number of words in the corpus, and allows us to compare language model performances on different corpus.

\section*{Acknowledgment}
This research is supported by Ministry of Culture, Sports and Tourism and Korea Creative Content Agency (Project Number: R2020040298).




\bibliographystyle{model5-names}\biboptions{authoryear}

\bibliography{reference}





\end{document}